\documentclass{article}

\usepackage[final]{nips_2018}
\usepackage{graphicx}
\usepackage{subcaption}
\usepackage{floatrow}
\newfloatcommand{capbtabbox}{table}[][\FBwidth]
\usepackage{blindtext}

\usepackage[utf8]{inputenc} 
\usepackage[T1]{fontenc}    
\usepackage{hyperref}       
\usepackage{url}            
\usepackage{booktabs}       
\usepackage{amsfonts}       
\usepackage{nicefrac}       
\usepackage{microtype}      

\title{Early Stratification of Patients at Risk for Postoperative Complications after Elective Colectomy}

\author{Wen Wang$^1$, Rema Padman$^1$, Nirav Shah$^2$\\
$^1$Heinz College of Information Systems and Public Policy, Carnegie Mellon University\\ $^2$NorthShore University HealthSystem and University of Chicago Pritzker School of Medicine \\
wenw3@andrew.cmu.edu, rpadman@cmu.edu, NShah2@northshore.org\\}

\begin{document}

\maketitle

\begin{abstract}
Stratifying patients at risk for postoperative complications may facilitate timely and accurate workups and reduce the burden of adverse events on patients and the health system. Currently, a widely-used surgical risk calculator created by the American College of Surgeons, NSQIP, uses 21 preoperative covariates to assess risk of postoperative complications, but lacks dynamic, real-time capabilities to accommodate postoperative information. We propose a new Hidden Markov Model sequence classifier for analyzing patients' postoperative temperature sequences that incorporates their time-invariant characteristics in both transition probability and initial state probability in order to develop a postoperative "real-time" complication detector. Data from elective Colectomy surgery indicate that our method has improved classification performance compared to  8 other machine learning classifiers when using the full temperature sequence associated with the patients' length of stay. Additionally, within 44 hours after surgery,  the performance of the model is close to that of full-length temperature sequence.
\end{abstract}

\section{Introduction}

Postoperative complications are a critical driver of healthcare expenditures. Moreover, postoperative patients without complications who are worked up unnecessarily incur significant morbidity and high costs. Early identification of patients at risk for postoperative complications, leading to timely and accurate workups, may reduce adverse events and readmission rates. However, current widely used predictive tools lack dynamic, real-time capabilities[1]. The American College of Surgeons (ACS) NSQIP is a national, validated, outcome-based, risk-adjusted, and peer-controlled program for measurement, benchmarking and enhancement of the quality of surgical care[3]. They create a Universal Surgical Risk Calculator, which uses 21 preoperative covariates to calculate risk of postoperative complications and has been helpful for discussing risk with patients and obtaining informed consent prior to surgery. However, once the surgery is completed, incorporating some surgery-related features and critical postoperative measurements, such as evolving temeprature sequences, are needed to identify postoperative complication risk in real-time in order to improve early diagnosis and avoid unnecessary testing and related adverse events.

Building on this knowledge, we propose a new Hidden Markov Model (HMM) sequence classifier of postoperative temperature sequences by incorporating patients' time-invariant characteristics in both transition probability and initial state probability for early detection of complications. A significant advantages of HMM compared to the traditional sequence classification method,which either transforms a sequence into a feature vector or sequence distance based classification method, is that HMM can model the dependence among elements in sequences[8][9][10], and the flexibility of the HMM structure allows us to incorporate prior knowledge in different ways[11]. The other advantage of HMM is that it can deal with variable length sequences. The probabilistic graphical model representation of the HMM also has strong interpretability which is important in healthcare area, whereby, for example, we can plot the path of hidden states to generate insights for medical decision making[2][5]. 

Besides, early classification of patients at risk for complications, using time series measurements such as temperature, is critical in our postoperative setting. Monitoring and classifying measurement sequences as early as possible is desired. The challenge of early classification is to study the trade off between the earliness and the accuracy of classification[7]. We explore how the real-time classifier works by testing on different lengths of sequences from a postoperative period of 24 hours up to 72 hours. We observe a sharp increase in performance at 44 hours, after which the overall performance is close to that of the full length of stay sequence.

\section{HMM with Covariates Sequence Classifer}
Given patient's post-operative sequence of temperature$O_{1:T}$, and three sets of time-invariant covariates $z$ which is basic patient characteristics(age, gender), surgery characteristics( total surgery procedure time etc) and eight pre-operative comorbidities diagnosed on admission(Uncomplicated HTN,Metastatic Cancer,Arrhythmias etc), our HMM will have following elements:

1. $S_t$ is one of low, medium, high risk hidden states w.r.t. complication risk

2.$\pi_i(z) = P(S_1 = i|z) $ is the probability of state i at time t=1 with covariates z, here we use multinomial logit model, state 1 is base category. Here we incorporate gender, age, total surgery procedure time(hours), eight pre-operative comorbidities diagnosed on admission (solid tumor without metastasis ,uncomplicated hypertension,arrhythmia, fluid and electrolyte disorder ,valvular disease,liver disease,chronic pulmonary disease,uncomplicated diabetes)  . The inital probability for state 1 is $\pi_1(z) = \frac{1}{1 + \sum_{j =2}^3 exp\{\alpha_j + \textbf{z}'\gamma_j\}}$, for state 2 and 3 is  $ \pi_i(z) = \frac{exp\{\alpha_i + \textbf{z}'\gamma_i\}}{1 + \sum_{j =2}^3exp\{\alpha_j + \textbf{z}'\gamma_j\}}   \quad\quad \quad  i = 2, 3$

3. $a_{ij}(z) = P(S_{t+1} = j|S_t = i, z)$ provides the probability of a transition from state i to state j with a vector of covariates z State 1 works as base category.Here we incorporate gender, age and total surgery procedure time in all there transition models. So transition probability from a specific state i (i can be 1, 2 or 3) to 1 is  $a_{i1}(z) = \frac{1}{1 + \sum_{j =2}^3exp\{\beta_{ij} + \textbf{z}'\eta_{ij}\}}$,  transition probability from a specific state i to 2 or 3 is  $ a_{ij}(z) = \frac{exp\{\beta_{ij} + \textbf{z}'\eta_{ij}\}}{1 + \sum_{j =2}^3exp\{\beta_{ij} + \textbf{z}'\eta_{ij}\}}\quad\quad \quad  j = 2,3$

4. $b_j(O_t) = P(O_t|S_t = j)$ is emission probability that provides the conditional densities of observation $O$ associated with latent class $j$. $b_j(o_t) = \mathcal{N}(\mu_j, \sigma_j)$

 The parameters of a HMM can be represented as  $\lambda = (\theta_1, \theta_2, \theta_3)$ consisting of three subvectors with parameters for prior model, transition model and response model.

In order to do classification,  two HMMs are derived, one
for complication patients ($\lambda_C$) and one for the non-complication patients ($\lambda_{NC}$),	given the trained model parameters($\lambda_C$,$\lambda_{NC}$) by Baum-Welch Algorithm ,and a set new sequences the models have never seen, the P(O|$\lambda$),probability sequence generating by each model, can be effectively solved by Forward- Backward Algorithm. Combined with Baysian formula, we can get the posterior probability according to which we can assign label to the new sequence:
\begin{equation}
P(C| O) = \frac{P(O|\lambda_C)*P(C)}{P(O|\lambda_C)*P(C) + P(O|\lambda_{NC})*P(NC)}
\end{equation}

Where C refers to complication, NC refers to non-complication,so $P(NC|O) = 1- P(C| O)$, $P(C)$ this posterior probability can be a real-time risk score, let's say we have a 24 hours sequence, when the new observation available, we can add this observation to the 24 hours sequence, and get a new risk score of current time step. $P(NC)$ is prior probability which we can take from general health statistics from our colectomy surgery dataset
\section{Data}
The health system maintains an electronic data warehouse (EDW) that captures information entered into the electronic health record. This data includes inpatient and outpatient health information such as demographics, vitals, labs, microbiology, radiology, medications, comorbidities, surgeries and costs. For the current study, data were extracted for all patients who underwent elective colectomy surgery between January 1, 2007 and December 31, 2013, to establish postoperative temperature patterns. Complications data was extracted through manual chart reviews. 

The dataset includes 632 unique patients with 22,938 temperature measurements across their inpatient stay after surgery and also the covariates of patient characteristics . We use the maximum temperature within each 4 hour interval, which is the metric used by clinicians to track onset of complications in the patient. Missing values are imputed when there is just one missing value between two intervals, resulting in just 6\% missing values in the dataset. The average length of stay of patients are 7.6 days, so the study horizon was selected to 10 days, we cut off all the temperatures after 10 days for each patient. 24\% of patients develop any complication after surgery. Here, we only consider inpatient complications. If the patient develops any type of complication (infectious or non-infectious), the patient will be considered as having 'any' complication . Our dataset is slightly imbalanced. To cope with this,we use Random Over Sampling techiniques (randomly add redundancy to the data set by duplicating data points of the minority classes). Other over sampling techniques like SMOTE,ADASYN are not considered, because it will add  noise to the data. Thus, the datasets include variable length temperature time series  and time-invariant covariates.

\section{EMPIRICAL RESULTS}
\subsection{Model Inference}
According to Table 1,we expect the higher post-operative temperature to be correlated with the higher risk of complication. So we name the three hidden states as low, medium and high risk status. One point should be noted is that all the gaussian means and standard deviations for complication model is higher than non-complication model. In order to see the different post-operative hidden states pattern for complication patients and non-complication patients, we use the Viterbi algorithm to find the "optimal" state sequence and plot the percentage of patients in each state over time. As shown in Figure 1, complication patients states  fluctuate more compared to non-complication patients. The percentage of complication patients in medium risk status reach the highest point at 24 hours after surgery, then fluctuates sharply over the time; the high risk status group show sharp increase and reach the peak at 45 hours. However, the share of medium risk for non-complication patients converge at 24 hours or so, the share of non-complication patients in high risk status is less than 0.1 at 72 hours, the smooth and steady prevalence of patients may indicate that patient recover to normal physiological level after the surgery and do not develop any complications.

\begin{minipage}{0.5\textwidth}
\begin{tabular}{|c|c|c|c|}
\hline
Parameters &  State 1&  State 2 & State 3\\
\hline
Gaussian Mean& 97.793&98.582&99.813\\
Gaussian SD&0.434&0.435&1.092\\
\hline
\end{tabular}

\end{minipage}
\begin{minipage}{0.5\textwidth}
\begin{tabular}{|c|c|c|c|}
\hline
Parameters &  State 1&  State 2 & State 3\\
\hline
 Gaussian Mean& 97.724&98.497&99.371\\
     Gaussian SD& 0.432& 0.372&0.900\\
\hline
\end{tabular}

\end{minipage}
\captionof{table}{Estimated mean and standard deviation of emission distributions from each latent state. Left: complication model, right: non-complication model}
\begin{minipage}{0.5\textwidth}
\centering
\includegraphics[width=1\textwidth]{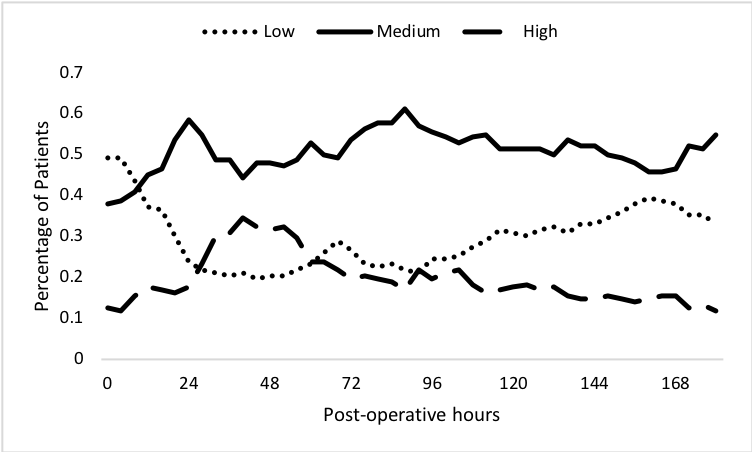}
\end{minipage}
\begin{minipage}{0.5\textwidth}
\centering
\includegraphics[width=1\textwidth]{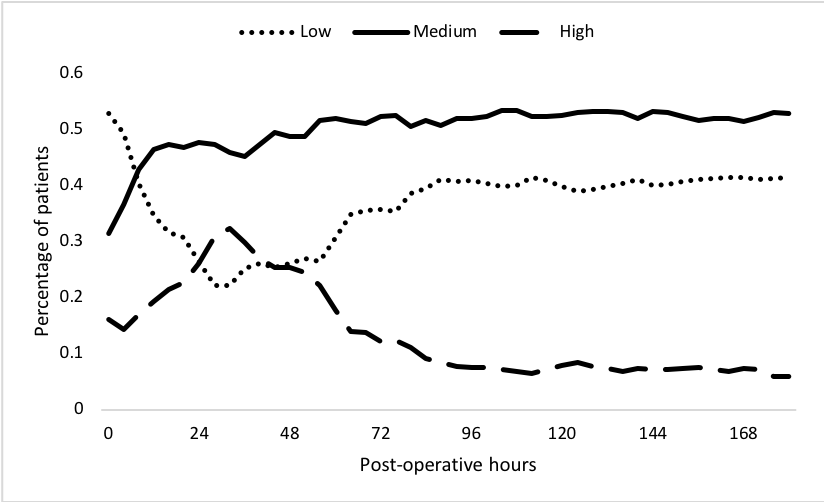}
\end{minipage}
\captionof{figure}{Percentage of patients in each state over time. Left:complication model, right: non-complication model}

\subsection{Classification Performance}

We first show the performance of our proposed classifier with full-length sequence  using 5 fold cross validation. We also show the performance of 8 other baseline classifiers: traditional machine learning classifier (Naive Bayes, SVM, random forest etc), ensemble learning classifier(Adaboost, bagging), and HMM sequence classifier without covariates in Table 2. Each baseline except HMM sequence classifier without covariates is tested with two sets of inputs: raw time series and time-invariant covariates. For raw time series, we use the first day sequence and last day sequence. This provides classifiers with temporal information about changes in patient state from end of the surgery to discharge or end of  our study period (10 days) within a fixed-size input, as required by all baseline methods. HMM without covariates classifier is tested with variable-length full-length sequences, which is exactly same as our proposed model excepting the covariates.Performance metrics adapted for imbalanced data problems are less likely to suffer from imbalanced distributions.
Receiver Operating Characteristics (ROC).  $G_{means}$,
and F-measure($F_m$) are used to compare the performance. $G_{Means} = \sqrt{\frac{TP}{TP+FN}*\frac{TN}{FP+TN}}$
$F-score = \frac{2(\frac{TP}{TP+FP}*\frac{TP}{TP+FN})}{\frac{TP}{TP+FP}+\frac{TP}{TP+FN}}$.Here use the Area Under ROC Curve (AUC) as a scalar measure instead of ROC curve. All results are average of 5 test results. We observe that our proposed model performs the best in comparison to decision tree, SVM and KNN. Naive Bayes, which is also widely used in sequence classification, assumes that features in the sequences are independent of each other. Its performance is worse than the HMM , becasue HMM can capture the dependency among elements.

\begin{minipage}{0.5\textwidth}
\centering
\includegraphics[width=1\textwidth]{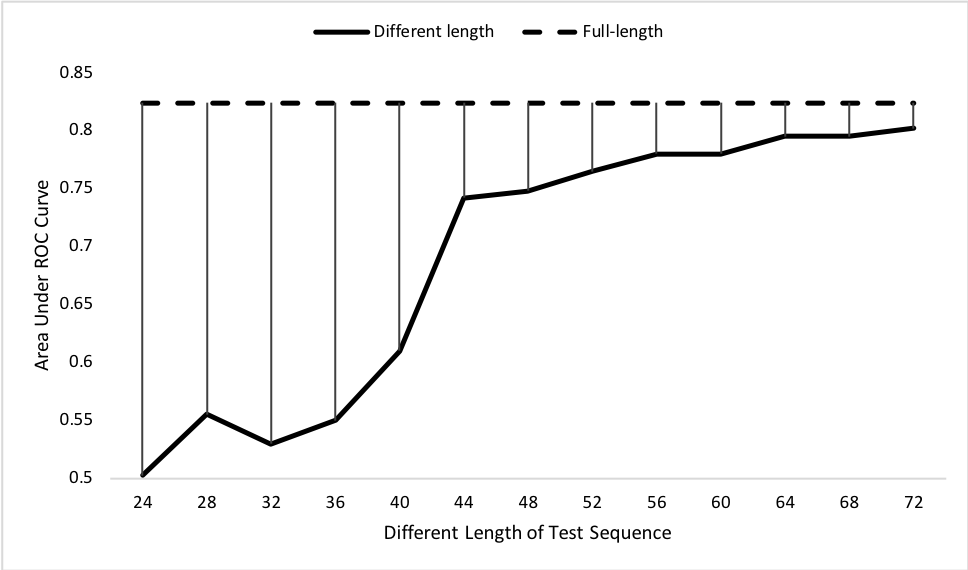}
\captionof{figure}{Early classification performance}
\end{minipage}
\begin{minipage}{0.4\textwidth}
\centering
\captionsetup{type=table} 
\begin{tabular}{|c|c|c|c|} \hline
Model & AUC& F-Score& $G_{Means}$ \\
\hline
Naive Bayes&0.763&0.205&0.528\\
SVM&0.669&0.294&0.541\\
Random forest&0.796&0.502&0.702\\
Adaboost&0.777&0.358&0.691\\
Bagging&0.757&0.372&0.544\\
KNN & 0.684 &0.451&0.679\\
HMM baseline&0.744&0.451&0.679\\
Decision Tree&0.573&0.113&0.267\\
\textbf{HMM with cov}&\textbf{0.824}&\textbf{0.625}&\textbf{0.783}\\
	\hline
  \end{tabular}
\caption{Classification performance compared with eight baseline methods}\label{tab: table-label}
\end{minipage}

\subsection{Early Classification}
In reality, the value of temperature sequence is received in time stamped ascending order. In our complication prediction setting, monitoring and classifying sequences as early as possible is desired. Xing et al. [62] point out the challenge of early classification is to study the trade off between the earliness and the accuracy of classification.
Therefore, we test the performance by using the different length of sequence from post-operative 24 hours sequences to 72 hours sequences. We use 5-fold cross validation here, train on the full-length sequence, but test on the different length sequence. We don't need to train multiple models for different length because HMM can deal with variable-length sequence.  Results show that performance before 40 hours is bad. There is a sharp increase at 44 hours. Performance of 44 hours onward is pretty close to that of full-length sequences according to AUC results in the Figure 2

\section{Conclusion}
In this study, in order to develop a post-operative "real-time" complication detector, we propose a new Hidden Markov Model sequence classifier by incorporating patients’ time-invariant characteristics in both transition probability and initial state probability. We demonstrate the classification performance of our proposed model by testing on 24-72 hours sequences as well as the full-length sequence (over the entire duration).  In  future work, we will incorporate additional vital sign sequences in our model (white blood cell counts, heart beat etc.), as well as test the addition of the vital signs one by one, to get a real-time score sequence (posterior complication probability), such that when the score is higher than a chosen threshold, an alert is generated.

\section*{Acknowledgements}
We are grateful to Dr. J. Grant and Dr. R. Parikh for inputs on the risk stratification problem and extraction of complications data. 
This study was approved by the Institutional Review Boards at Carnegie Mellon University and NorthShore University HealthSystem.

\section*{References}

\small

[1] Bilimoria, K. Y., Liu, Y., Paruch, J. L., Zhou, L., Kmiecik, T. E., Ko, C. Y., and
Cohen, M. E. Development and evaluation of the universal acs nsqip surgical risk
calculator: a decision aid and informed consent tool for patients and surgeons.
{\it Journal of the American College of Surgeons 217}, 5 (2013), 833–842.

[2] Bilmes, J. A., et al. A gentle tutorial of the em algorithm and its application to
parameter estimation for gaussian mixture and hidden markov models. {\it International
Computer Science Institute 4}, 510 (1998), 126.

[3] Cohen, M. E., Bilimoria, K. Y., Ko, C. Y., and Hall, B. L. Development of an
american college of surgeons national surgery quality improvement program:
morbidity and mortality risk calculator for colorectal surgery. {\it Journal of the
American College of Surgeons 208}, 6 (2009), 1009–1016.

[4] Galicier, C., and Richet, H. A prospective study of postoperative fever in
a general surgery department. {\it Infection Control \& Hospital Epidemiology 6}, 12
(1985), 487–490.

[5] Rabiner, L. R. A tutorial on hidden markov models and selected applications in
speech recognition.{\it Proceedings of the IEEE 77}, 2 (1989), 257–286.

[6] Shah, N., Dhingra, A., Ridgway, J. P., Padman, R., and Robicsek, A. What
is a normal postoperative temperature? group based trajectory modeling in
postoperative knee arthroplasty patients in a large health system. {\it In Open Forum
Infectious Diseases (2015)}, vol. 2, Infectious Diseases Society of America, p. 368.

[7] Z. Xing, J. Pei, G. Dong, and P. S. Yu. Mining sequence
classifiers for early prediction. In {\it  SDM'08: Proceedings
of the 2008 SIAM international conference on data mining}, pages 644 2008.

[8] Al-Ani, Tarik, et al. "Using hidden Markov models for sleep disordered breathing identification." {\it Simulation Modelling Practice and Theory} 12.2 (2004): 117-128.

[9] Georgoulas, George G., et al. "Classification of fetal heart rate during labour using hidden Markov models." Neural Networks, 2004.{\it  Proceedings. 2004 IEEE International Joint Conference on. Vol. 3}. IEEE, 2004.

[10] Xing, Zhengzheng, Jian Pei, and Eamonn Keogh. "A brief survey on sequence classification."{\it ACM Sigkdd Explorations Newsletter 12.1 (2010)}: 40-48.

[11] Yi, Kwan, and Jamshid Beheshti. "A hidden Markov model-based text classification of medical documents." {\it Journal of Information Science 35.1} (2009): 67-81.
\end{document}